\documentclass{article}

\usepackage{microtype}
\usepackage{graphicx}
\usepackage{subfigure}
\usepackage{booktabs} 

\usepackage{hyperref}

\usepackage[accepted]{mlsys2021}

\mlsystitlerunning{CREST: Effectively Compacting a Datastore for Retrieval-Based Speculative Decoding}

\begin{document}

\twocolumn[
\mlsystitle{CREST: Effectively Compacting a Datastore for \\Retrieval-Based Speculative Decoding}



\mlsyssetsymbol{equal}{*}

\begin{mlsysauthorlist}
\mlsysauthor{Sophia Ho}{equal,cmu}
\mlsysauthor{Jinsol Park}{equal,cmu}
\mlsysauthor{Patrick Wang}{equal,cmu}
\end{mlsysauthorlist}

\mlsysaffiliation{cmu}{School of Computer Science, Carnegie Mellon University}

\mlsyscorrespondingauthor{Sophia Ho}{sho2@cs.cmu.edu}
\mlsyscorrespondingauthor{Jinsol Park}{jinsolp@cs.cmu.edu}
\mlsyscorrespondingauthor{Patrick Wang}{phw2@cs.cmu.edu}

\mlsyskeywords{Machine Learning, MLSys, Speculative Decoding}

\vskip 0.3in

\begin{abstract}
We present CREST (Compact Retrieval-Based Speculative Decoding), a redesign of REST that allows it to be effectively ``compacted". REST is a drafting technique for speculative decoding based on \textit{retrieving} exact n-gram matches of the most recent n tokens generated by the target LLM from a datastore. The key idea of CREST is to only store a \textit{subset} of the smallest and most common n-grams in the datastore with the hope of achieving comparable performance with less storage space. We found that storing a subset of n-grams both reduces storage space and \textit{improves} performance. CREST matches REST's accepted token length with 10.6-13.5x less storage space and achieves a 16.5-17.1\% higher acceptance length than REST using the same storage space on the HumanEval and MT Bench benchmarks.

\end{abstract}
]



\printAffiliationsAndNotice{\mlsysEqualContribution} 

\section{Introduction}
\label{sec:intro}
Recently, Speculative Decoding has gained traction for accelerating LLM inference by employing a small draft model to reduce inference latency without sacrificing the quality of the outcome sequence \cite{spector2023accelerating, chen2023accelerating, yan2024decoding}. The small draft model is used to ``draft" tokens which are sent to the LLM for verification. If the LLM verifies the predictions from the small model are correct, then those draft tokens are accepted, otherwise, they are regenerated by the LLM. This approach is efficient because the large model verification step can be done in parallel. However, the success of this strategy is heavily dependent on the quality of the small draft model and this often requires training a draft model for specific language tasks. This introduces additional complexity and overhead in the training and deployment of a second language model.

REST \cite{he2023rest}, which stands for retrieval-based speculative decoding, takes a different approach. Instead of using a draft model, REST uses a static datastore constructed from a pre-training dataset. REST takes the most recent $n$ tokens generated by the LLM and attempts to find exact matches in the datastore. It then combines the sequences of tokens after the exact matches into a token tree to pass to the LLM. The key benefit of REST is that it does not require any ML components, allowing it to work out-of-the-box with any LLM.

Although REST can achieve a high draft token acceptance rate, the static nature of the datastore introduces a new challenge regarding storage space. REST stores the entire text of a pre-training dataset as-is, and the way to improve the accuracy of REST is to simply append more text to the datastore. However, this grows the size of the datastore unboundedly, motivating the need for a method to ``compact" a datastore. Due to the way REST is designed (see \autoref{sec:rest_datastore_problems}), attempting to compact the datastore by removing specific tokens in the middle of the dataset is highly disruptive and not a feasible strategy.

Our method, which we have named CREST (Compact Retrieval-Based Speculative Decoding) addresses these flaws by redesigning REST such that it is possible to remove select tokens. At a high level, we ``decouple" n-grams from each other in the datastore such that tokens do not simultaneously serve as parts of multiple n-grams. With this redesign, it is possible to intelligently select a \textit{subset} of n-grams to keep, allowing the datastore to be effectively compacted. We tried a strategy of only keeping the \textit{smallest} and \textit{most common} n-grams. We found that this reduced storage space and, surprisingly, \textit{increased} our accepted length.


We summarize our contributions below:
\begin{enumerate}
\item We \textbf{redesign the underlying structure} of the REST datastore to "decouple" n-grams from each other.
\item We develop a compaction strategy that has better compaction behavior than REST and \textbf{outperforms REST at all datastore sizes}.
\item In addition to performing better at smaller sizes, CREST can achieve up to \textbf{17.1\% higher accepted length} than REST's peak accepted length.
\item When comparing ideal implementations of REST and CREST, our redesigned datastore \textbf{reduces drafting time complexity from $O(\log n)$ to $O(1)$}.
\end{enumerate}



\section{Related Work}
\label{sec:related}

\subsection{Speculative Decoding}
Speculative Decoding can be conceptualized as a decoding strategy known as Draft-then-Verify, where the process involves generating multiple potential tokens at each step and then verifying them concurrently against the target Language Model (LLM) to expedite inference \cite{xia2024unlocking-survey}.

Speculative Decoding started from Blockwise Decoding introduced in \citet{stern2018blockwise}, which enhanced the Transformer decoder with additional feed-forward neural network (FFNN) heads. This augmentation facilitates the simultaneous generation of multiple tokens in each step, subsequently validated in parallel by the original LLM to ensure coherence with its outputs.

Subsequent research employed smaller language models as drafting agents. For instance, \citet{xia2023speculative} employs a specialized non-autoregressive transformer as an independent drafter. Similarly, other works \cite{leviathan2023fast, chen2023accelerating} utilize smaller versions of LLMs as drafting models, enabling speculative decoding without the need for additional fine-tuning.

Some approaches choose to utilize the target LLM itself instead of introducing additional drafting models. Medusa \cite{cai2024medusa} integrates FFNN heads into the transformer decoder, enabling parallel token generation per step.
Other works \cite{yang2023predictive, zhang2023draft} take a more systemic approach, and choose to use additional subprocesses or adaptively skip several intermediate layers for efficiency.

Alternatively, certain methods focus on enhancing parallel decoding performance without altering the model architecture. \citep{monea2023pass} introduces learnable tokens, and fine-tune their new embeddings on a small training dataset to improve parallel decoding efficiency.

\subsection{Retrieval-Based Speculative Decoding (REST)}
In contrast to other approaches, REST utilizes an existing text dataset to facilitate speculative decoding. It achieves this by retrieving n-grams from the datastore that match the latest $n$ tokens of the current context. Specifically, REST operates on a pre-constructed datastore comprising pairs of (context, continuation), employing an exact-match method for fast retrieval. After gathering retrieved results, REST combines the continuations into a token tree to pass to the LLM for verification.

A key advantage of REST is that it does not use ML components to draft tokens, thus eliminating the need for any form of training or serving. In contrast to previous methods that rely on a draft language model for speculative decoding, REST can be integrated out-of-the-box with any target LLM, without requiring additional training. 


\section{Problem}
\label{sec:problem}

Our approach is motivated by the difficulty of effectively ``compacting" the REST datastore while retaining as much performance as possible. Such compaction is important to avoid the size of the datastore growing unboundedly as we use larger and larger pre-training datasets to build the datastore. We first give detailed background on the REST datastore's design, discuss the problems we noticed with it, and give a high level motivation of our proposed solution.

\subsection{REST Datastore Background}
\label{sec:rest_datastore_background}
\textbf{Datasets and flattened datasets:} In the context of REST, a \underline{dataset} is an \textit{unordered} set of conversations. A \underline{conversation} refers to an \textit{ordered} sequence of alternating prompts and responses. A \underline{prompt} is simply an ordered sequence of tokens generated by a human while a \underline{response} is an ordered sequence of tokens generated by an LLM.

While a dataset is a complex nested data structure, it can be \underline{flattened} into a single ordered sequence of tokens. First, each conversation is \underline{flattened} by concatenating the tokens of the prompts and responses together. Then, the conversations are put in an arbitrary order and their flattened representations are concatenated. The final representation is called a \underline{flattened dataset}. For the rest of the paper, when we use the word ``dataset", we are referring to a flattened dataset.

\textbf{REST data structures:} The REST datastore takes a \underline{flattened dataset} and splits the conversations of the dataset into \underline{chunks} of token sequences. The order of tokens within a chunk is maintained from the flattened dataset, but the order of different chunks in the REST datastore is arbitrary. In the current implementation of REST, each chunk is 512MB (except for the last chunk, which may be smaller).

\begin{figure}[t]
\centering
\includegraphics[width=0.5\textwidth]{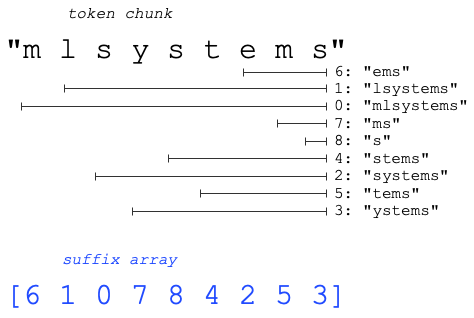}
\caption{An example suffix array using the token chunk \texttt{"mlsystems"}, where each character is a token. Note that in REST, each token is a 32-bit integer and not a character. We made each token a character in this example only for ease of explanation.}
\label{fig:suffix_array}
\end{figure}

Then, REST creates an auxiliary index structure over each chunk called a \underline{suffix array} \cite{suffix-array}. This structure is a list of integers where each integer refers to a position in the 512MB chunk of tokens. Each integer conceptually represents a \underline{suffix} of the 512MB chunk, which refers to the sequence of tokens from that integer position in the chunk until the end of the chunk. The integers in the suffix array are ordered such that the suffixes represented by the integers are in alphabetical order. \autoref{fig:suffix_array} shows an example of a suffix array. Note that while the figure shows a suffix array where each token is a character, in REST, each token is a 32-bit integer.

\textbf{REST algorithms:} The REST datastore consists of two main algorithms: one to construct the datastore and one to query the datastore. Constructing the datastore is done with \hyperlink{https://github.com/IlyaGrebnov/libsais}{\texttt{libsais}}, so we will not discuss it here.

\begin{figure}[t]
\centering
\includegraphics[width=0.5\textwidth]{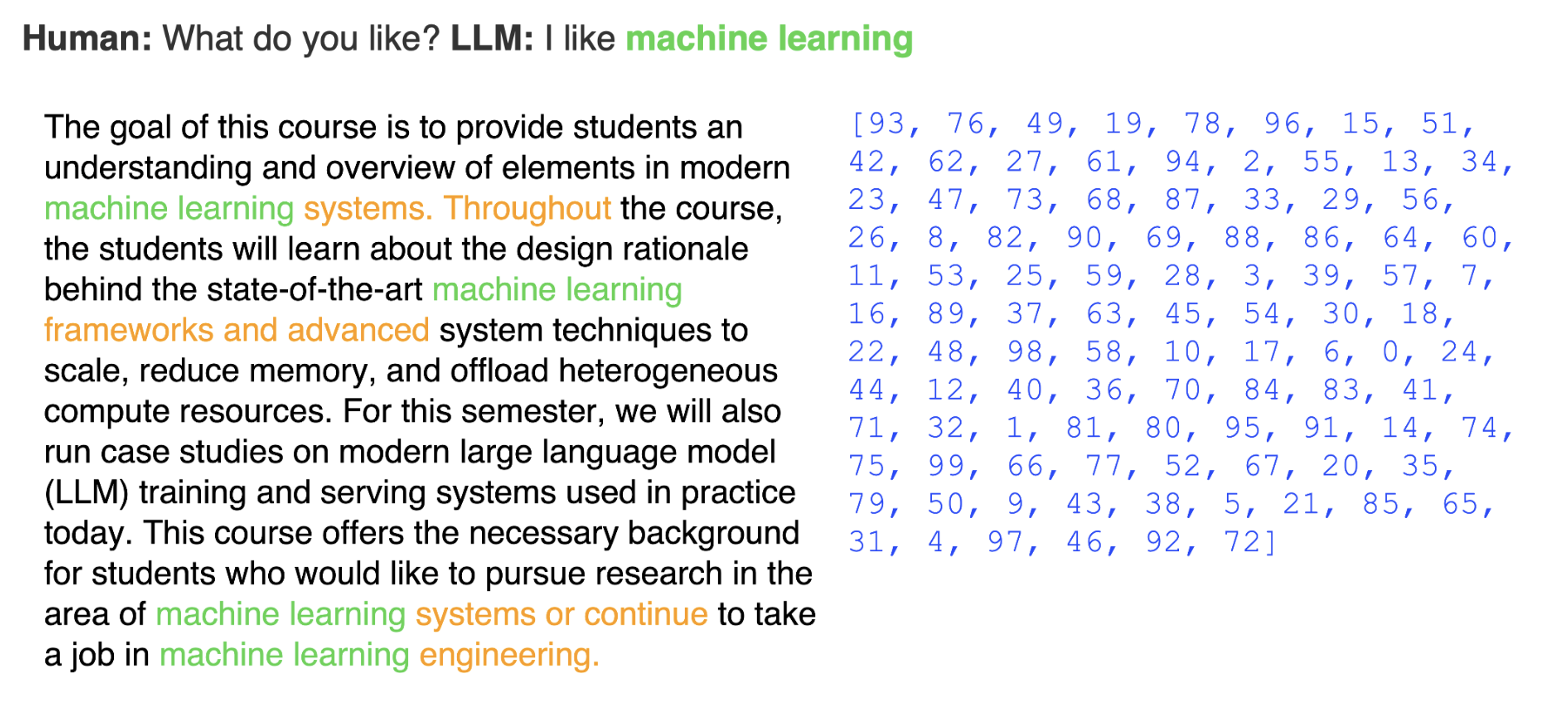}
\caption{An example of REST's drafting process. The context is the latest 2 generated tokens, or ``machine learning". The exact matches of the context are highlighted in green in the datastore. The continuations are highlighted in orange.}
\label{fig:rest_drafting}
\end{figure}

When querying, REST first tries to find 16-grams in the datastore that exactly match the \underline{context}, which refers to the latest 16 generated tokens (e.g. the ``suffix" of what the LLM has generated so far). In \autoref{fig:rest_drafting}, this context is ``machine learning". If it finds at least one \underline{matching context}, it will build a \underline{token tree} using the \underline{continuations} (sequences of tokens) immediately following all matching contexts. The continuations are the tokens highlighted in orange in \autoref{fig:rest_drafting}. If it has found at least one exact match, it combines the continuations of the exact matches it found into a token tree. If it finds zero exact matches, it will try to find at least one exact 15-gram match. It continues doing this down to 2-grams, at which point it stops.

To find an exact match efficiently, REST performs a binary search over the suffix arrays of each chunk in parallel. One unintuitive detail is that there are two different notions of ``suffixes" at play. The suffix array conceptually contains suffixes that stretch from a token in the chunk to the end of the chunk, meaning their lengths could be extremely long. However, the ``suffix" of what the LLM has generated so far is at most 16 tokens in length. Confusingly, the ``suffix" of what the LLM has generated so far is meant to match the \textit{prefix} of the ``suffixes" in the suffix array.

\subsection{REST Datastore Problems}
\label{sec:rest_datastore_problems}
\textbf{Density/Layering:} With the suffix array, REST can store its datastore in an extremely dense fashion. N-grams of different sizes are ``layered" in place on top of each other, and the same tokens serve simultaneously as both contexts and continuations in different situations. While such density is impressive, it does make it difficult to compact the REST datastore by selectively removing tokens. Fundamentally, this is because removing a single token in the REST datastore partitions the chunk into two pieces, eliminating about $2m-1$ contexts and $2l-1$ continuations where $m$ is the size of a matched context while $l$ is the length of the continuation\footnote{Note that if the removed token was already near the edge of the chunk, it might not remove as many contexts or continuations. Also, note that REST matches contexts of sizes 2 through $m$, so we are actually eliminating more than $2m-1$ contexts.}. The most straightforward way to increase drafting accuracy in REST is to add more data to the datastore. However, this causes the datastore's size to grow unboundedly because there is no way to subsequently ``compact" the datastore.


\textbf{Time complexity:} REST's search algorithm has a time complexity of $O(c\log n_c)$, where $c$ is the number of chunks and $n_c$ is the average number of tokens in a chunk. Currently, each chunk is of a constant size, meaning that REST's overall time complexity is $O(n)$ where $n$ is the total number of tokens in the datastore. Even with an ideal implementation of REST where $c=1$, the best achievable time complexity in REST is $O(\log n)$.

\subsection{Compact-REST (CREST) Motivation}
\label{sec:crest_motivation}
In CREST, we aim to solve both of the problems mentioned in \autoref{sec:rest_datastore_problems} at the same time. Instead of densely ``layering" all contexts and continuations on top of each other, our idea was to ``decouple" the contexts and continuations into a dictionary, where contexts would map directly to token trees built from continuations and be completely separate from one another. Decoupling all the contexts from each other yields a large \textit{initial} increase in datastore size but opens up the opportunity for compacting the datastore by choosing only a \textit{subset} of the contexts. Additionally, it resolves the time complexity problem because all token trees are precomputed and only need to be looked up, which can be done in $O(1)$ time.

\begin{figure}[t]
\centering
\includegraphics[width=0.5\textwidth]{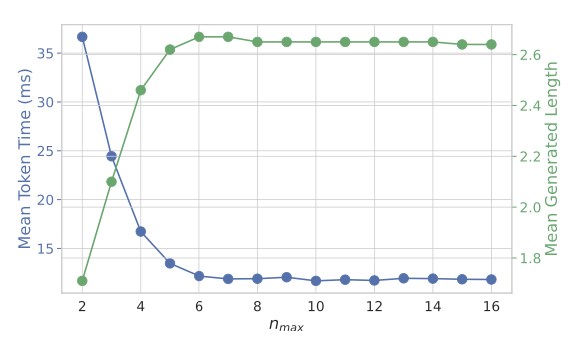}
\caption{Generation speed and average accepted length of REST with different starting context lengths. The graph is lifted directly from the REST paper. The settings are CodeLlama 7B with greedy sampling on HumanEval.}
\label{fig:rest_n-grams_ablation}
\end{figure}

We had a few pieces of evidence for why we believed that it would be possible to intelligently select a subset of n-grams to keep in the datastore. First, from the REST paper, we found that most of their speculation performance could be achieved by matching n-grams of $n\leq 5$ (see \autoref{fig:rest_n-grams_ablation}) even though they tried to match n-grams up to $n\leq 16$. We also hypothesized that a small percentage of the most common n-grams would be responsible for most of the speculation performance of REST. Our project's goal was to explore what would happen if you only stored the \textit{smallest} and \textit{most common} n-grams.

\begin{figure}[t]
\centering
\includegraphics[width=0.5\textwidth]{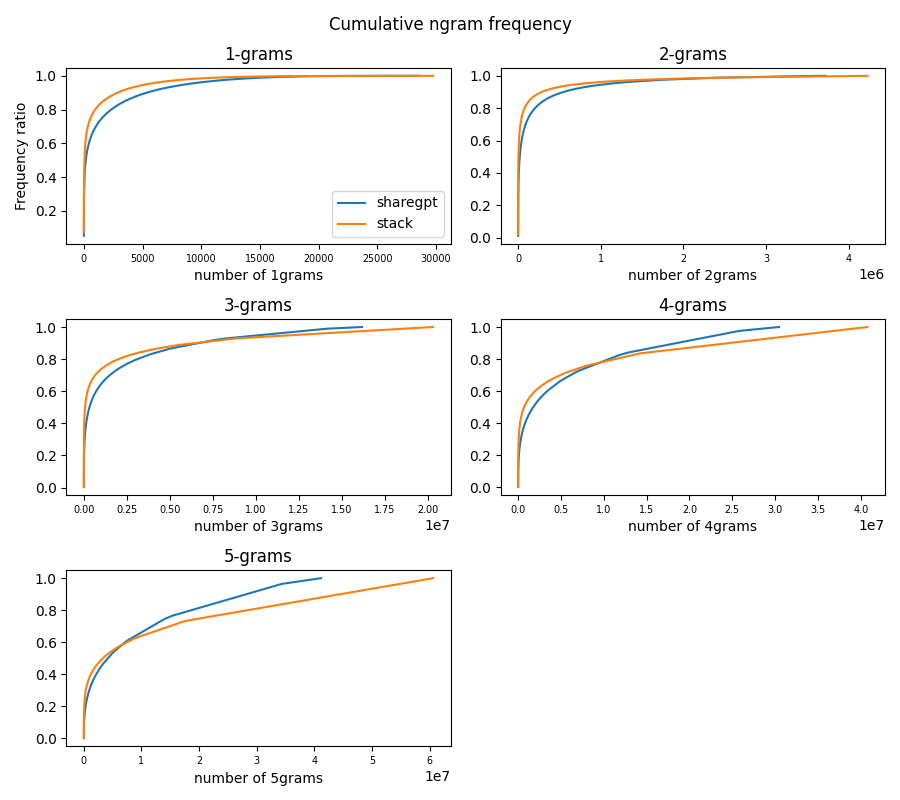}
\caption{Frequency analysis for n-grams up to 5 for the ShareGPT and The Stack dataset.}
\label{fig:n-gram_frequencey_merged}
\end{figure}

We performed a quick sanity check to see if this high-level direction was feasible. REST used two datasets, and their datastores for both were $\sim$1GB From \autoref{tab:unique-n-grams}, we found that both of our datasets had $\sim$100M unique n-grams of size $n\leq 5$. Each n-gram would store a token tree of 64 tokens, which is about 256 bytes as our tokens are 32-bit integers. Thus, storing all unique-n-grams would take $\sim$25GB. This means that we could store $\sim$4\% of the most common n-grams without exceeding the size of REST's datastore. We graphed the cumulative frequencies of n-grams in \autoref{fig:n-gram_frequencey_merged} and found that the distribution was very tail-heavy. Thus, storing the top $\sim$4\% of n-grams seemed feasible.

\begin{table}[t]
\caption{Number of unique n-grams for different $n$ in two datasets.}
\label{tab:unique-n-grams}
\vskip 0.15in
\begin{center}
\begin{small}
\begin{sc}
\begin{tabular}{lcccr}
\toprule
$n$ & ShareGPT & Stack \\
\midrule
1 & 28,598 & 29,777 \\
2 & 3,720,951 & 4,228,141 \\
3 & 16,178,050 & 20,320,491 \\
4 & 30,462,651 & 40,790,084 \\
5 & 41,164,874 & 60,629,482 \\
1-5 & 91,555,124 & 125,997,975 \\
\bottomrule
\end{tabular}
\end{sc}
\end{small}
\end{center}
\vskip -0.1in
\end{table}

\section{Methods}
\label{sec:methods}

\subsection{System Overview}
\begin{figure}[t]
\centering
\includegraphics[width=0.5\textwidth]{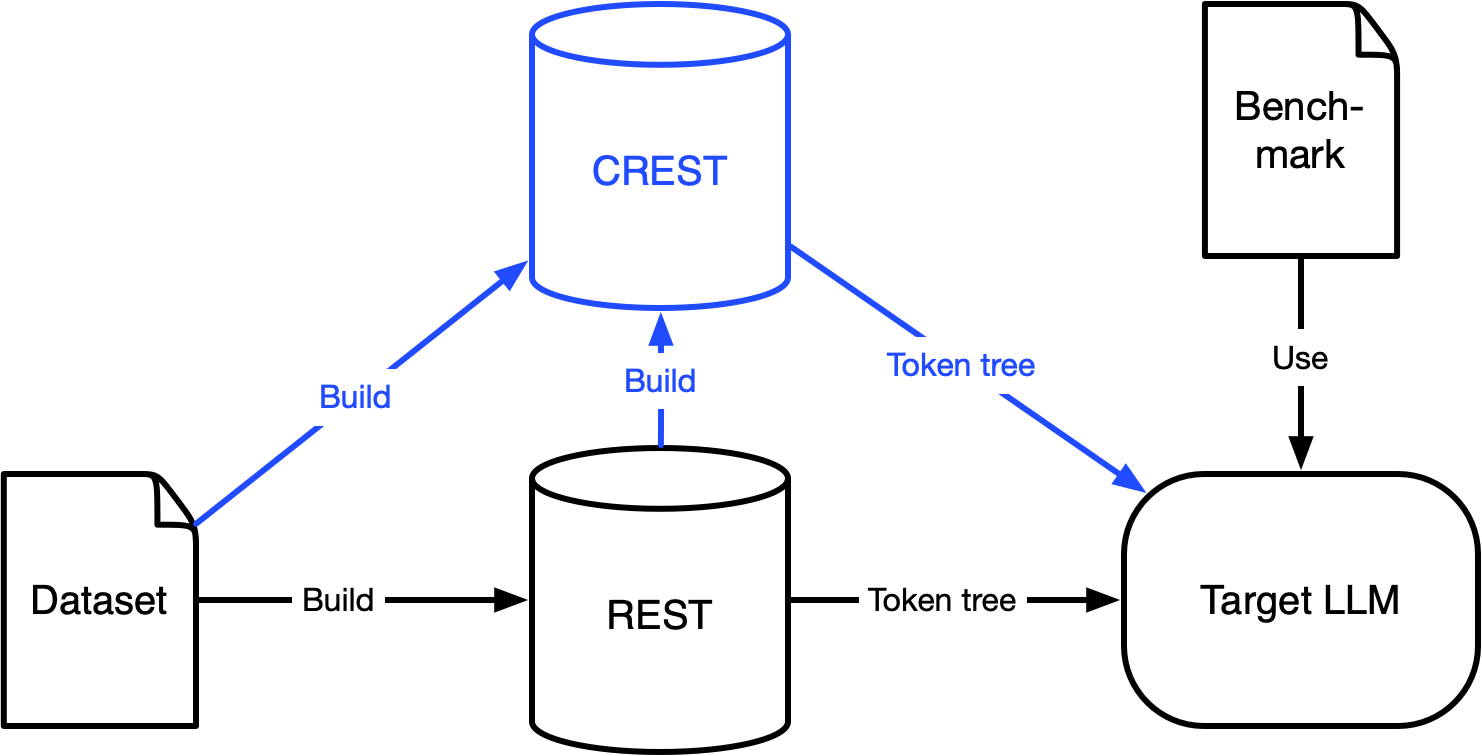}
\caption{An overview of the CREST system. Our contributions are highlighted in blue. Note that arrows are highlighted in addition to components because many challenges we faced involved integration issues, not just the core CREST components.}
\label{fig:system_overview}
\end{figure}

\autoref{fig:system_overview} shows the different components of the CREST system. Our contributions are highlighted in blue. Note that arrows are highlighted in addition to components because many challenges we faced involved integration issues, not just the core CREST components.

The \underline{dataset} contains past conversations between a human and an LLM. The \underline{REST datastore} is built by flattening the dataset and building a suffix array over it (see \autoref{sec:rest_datastore_background}). Next, the \underline{CREST datastore} is built by analyzing the dataset to find the \textit{smallest} and \textit{most common} n-grams and then querying the REST datastore with these n-grams. Both of these datastores are then used to draft token trees to pass to a \underline{Target LLM} for verification. Finally, the accepted length of both datastores is compared under a \underline{benchmark}.

\subsection{Building the CREST Datastore}
\label{sec:crest_datastore_building}
The CREST datastore is a hash map that maps from n-grams to token trees. To be scalable, this hash map must be ``disk-native". By ``disk-native", we mean that you can directly traverse the file on disk to find matching n-grams (i.e. it requires $O(1)$ memory). Additionally, a key feature of CREST is that the n-grams it stores are a \textit{subset} of the n-grams present in the dataset.

\textbf{Building an efficient disk-native datastore:} Our original prototype was a Python dictionary which we serialized to disk with the \texttt{pickle} library. To make this approach disk-native, we switched to using Postgres. Instead of pickling the entire dictionary, we pickled each token tree and stored them as blobs in Postgres. Using an index on n-grams, we can support either $O(\log n)$ or $O(1)$ lookups for exact n-gram matches depending on whether a tree-based or hash-based index is used\footnote{We use a tree-based index because Postgres requires that all primary keys have a tree-based index. However, it is easy to imagine a database purpose-built for retrieval-based drafting that gets around this limitation.}. Additionally, we noticed that the token trees contained attention masks which had a large number of consecutive 0s and 1s, so we utilized the \texttt{lzma} Python library to compress these values before storing them in the database.

\textbf{Intelligently selecting a subset of n-grams:} There are two benefits to only storing a \textit{subset} of the n-grams in the CREST datastore. The first benefit is obvious: storing a subset of the n-grams takes up less storage space than storing the complete set of n-grams. The second benefit was surprising to us: storing a subset of the n-grams can lead to a \textit{higher} accepted length. The key reason behind this is that at any point in time during inference, there is more than one n-gram that matches the suffix of the currently generated text. By removing n-grams with ``low quality" token trees from the datastore, we bias our system to choose n-grams with ``higher quality" token trees on average.

There are many possible ways to select a subset of n-grams, but we investigated the idea of selecting the smallest and most common n-grams for this project. One important design decision we thought about was how to handle n-grams of different values of $n$. We chose to investigate both storing the most common $t$ n-grams of a single value of $n$ as well as storing the most common $t$ n-grams of all $n$ less than or equal to some "maximum size" $m$. In \autoref{sec:eval}, we found that the second method performed better than the first.

\section{Evaluation}
\label{sec:eval}
\subsection{Hardware and Benchmarks}
We conducted all our experiments using one NVIDIA L4 Tensor Core GPU and 8 vCPUs\footnote{Initially our experiment environment was on the PSC machines, however, we had many challenges during setup so we migrated our experiments to AWS.}. We built REST datastores using the \hyperlink{https://huggingface.co/datasets/Aeala/ShareGPT_Vicuna_unfiltered}{ShareGPT} and \hyperlink{https://huggingface.co/datasets/bigcode/the-stack}{The Stack} datasets, which come out to 1.1GB and 924MB of disk storage respectively. The ShareGPT dataset consists of 120675 training samples and The Stack dataset contains 90016 training samples. The draft tokens retrieved from the ShareGPT datastore were passed to Vicuna \cite{vicuna2023} and evaluated with the MT Bench \cite{zheng2023judging} benchmark. MT Bench is a multi-turn question set consisting of 80 multi-turn questions intended to mimic open-ended questions. The tokens drafted from The Stack datastore were sent to CodeLlama \cite{roziere2023code} and evaluated with the HumanEval \cite{chen2021evaluating} benchmark. The HumanEval dataset consists of 164 human-written coding problems and evaluates a model's ability to generate correct code from the docstrings. We used the 7B versions of Vicuna and CodeLlama and set the maximum number of tokens generated to 1024 and 512, respectively. We used REST's default hyperparameters: a maximum token tree size of 64, a maximum of 5000 matched contexts per search, and 10 token continuations after each matched context.


\subsection{Comparison with REST}
\label{sec:comparison-with-rest}
\begin{figure}[t]
\centering
\includegraphics[width=0.5\textwidth]{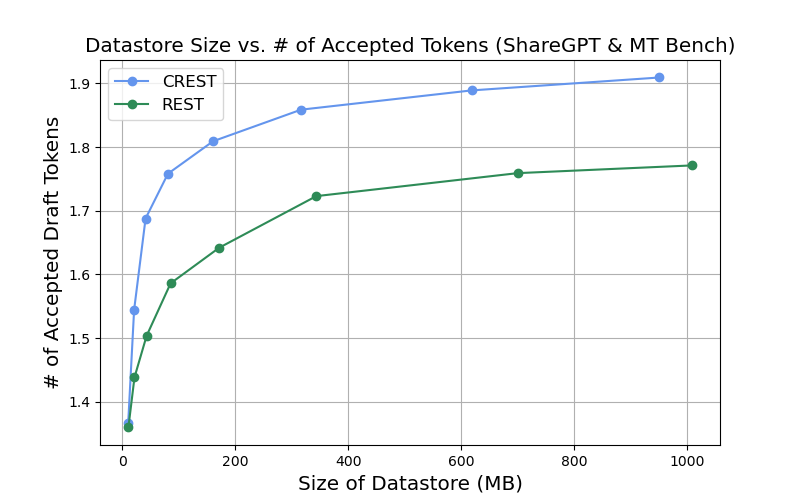}
\caption{Draft token acceptance length for each datastore size using data from ShareGPT and the MT Bench evaluation metric.}
\label{fig:rest_sharegpt_exp}
\end{figure}

\begin{figure}[t]
\centering
\includegraphics[width=0.5\textwidth]{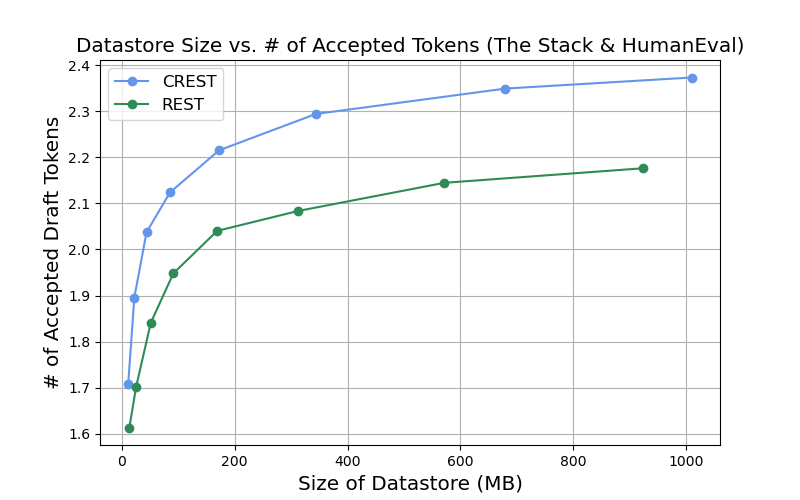}
\caption{Draft token acceptance length for each datastore size using the dataset The Stack and the HumanEval evaluation metric.}
\label{fig:rest_stack_exp}
\end{figure}
As our baseline, we constructed REST datastores using a random sample of 1\%, 2\%, 4\%, 8,\% 16\% 32\%, 64\%, and 100\% of the training samples for each of the two datasets. We constructed CREST datastores using our combined top-t strategy (see \autoref{sec:combining-top-t}). We used $n\leq 4$ for the ShareGPT CREST datastore and $n\leq 5$ for the Stack CREST datastore. We used values of $t$ which we estimated would yield CREST datastores of similar sizes to the REST datastores we built.

Overall, CREST outperforms REST at every datastore size on both MT Bench and HumanEval. \autoref{fig:rest_sharegpt_exp} demonstrates the performance of CREST compared with REST on the ShareGPT data evaluated on MT Bench. REST achieves a high of a 1.77 draft token acceptance length at the 100\% datastore size. CREST surpasses this metric and reaches a 1.9 draft token acceptance length. Both CREST and REST perform better with data from The Stack evaluated on HumanEval as seen in \autoref{fig:rest_stack_exp}. For HumanEval, REST attains a maximum accepted token length of 2.17, while CREST exceeds this with an accepted token length of 2.37.

Our initial goal was to \textit{sacrifice} performance to reduce storage size, so we were surprised to see that we had a higher accepted length than REST. We believe this is due to how the lookup algorithms of both REST and CREST start by looking for exact 16-gram matches and keep looking for smaller and smaller matches until they find at least one match in the datastore (see \autoref{sec:rest_datastore_background}). If all n-grams are kept, we could terminate early on an uncommon n-gram that has a ``lower quality" or smaller token tree.

While our token trees are compressed (see \autoref{sec:crest_datastore_building}), we notably \textit{do not} compress the REST datastore file when comparing disk usage. We believe this is the most fair comparison because you would otherwise need to uncompress the REST datastore before being able to search through it. On the other hand, our database is fully operational even while the token trees are compressed. Further, due to how densely the REST datastore is laid out (see \autoref{sec:rest_datastore_background}), we do not believe there are any trivial ways of compressing any parts of it while leaving it fully operational.

\subsection{Single $n$ Top $t$ Experiments}
\label{sec:single-n}
When it comes to selecting the most common n-grams, it is non-trivial to compare the frequencies of n-grams of different sizes (see \autoref{sec:crest_datastore_building}). This is because, given an n-gram $g$ of size $n$, any n-gram $g'$ of size $<n$ that appears within $g$ is guaranteed to show up at least as often as $g$. Given this, we first tried building CREST using the top $t$ most common n-grams for single values of $n$.

\begin{figure}[t]
\centering
\includegraphics[width=0.5\textwidth]{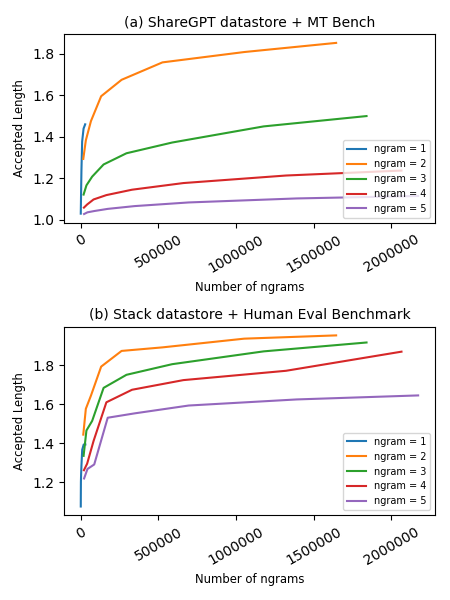}
\caption{Accepted length over the first 20 conversations across various datastore sizes and n-gram sizes when using only a single value of $n$. (a) shows the result of running 20 conversations from MT Bench, using a datastore from ShareGPT. (b) shows the result of running 20 prompts from Human Eval benchmark, using a datastore from The Stack.}
\label{fig:individual_n_exp}
\end{figure}

Our results are shown in \autoref{fig:individual_n_exp}. As the number of n-grams increases, the accepted length increases, which is expected. However, the rate at which the accepted length increases decreases, confirming our hypothesis that common n-grams are more important to have in the datastore than uncommon n-grams. On both the ShareGPT datastore and Stack datastores, $n=2$ gives the highest accepted length compared to any other individual value of $n$. However, the relative "drop-off" of increasing $n$ is steeper for ShareGPT than it is for Stack. One possible reason for this is that code data (which Stack contains) is more structured, causing larger n-grams to be more consistent in predicting the tokens that come after.

\subsection{Multiple $n$ Top $t$ Experiments}
\label{sec:combining-top-t}

\begin{figure}[t]
\centering
\includegraphics[width=0.5\textwidth]{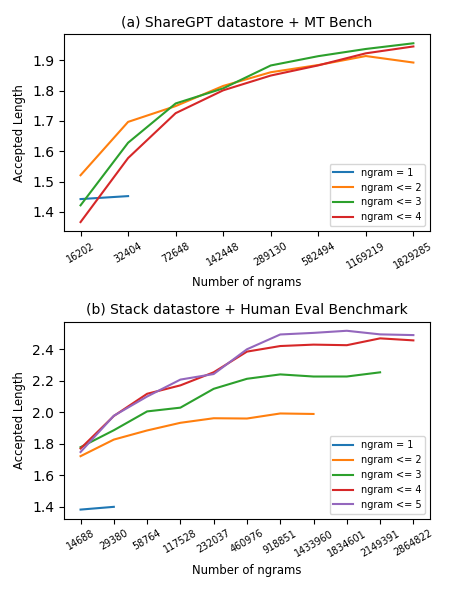}
\caption{Accepted length over the first 20 conversations across various datastore sizes and n-gram sizes when using all values of $n$ up to some maximum $m$. (a) shows the result of running 20 conversations from MT Bench, using a datastore from ShareGPT. (b) shows the result of running 20 prompts from Human Eval benchmark, using a datastore from The Stack.}
\label{fig:topN_exp}
\end{figure}

To examine the trend of accepted lengths across various datastore sizes and n-gram configurations, we conducted experiments on a subset of the entire benchmark dataset.
For each line "ngram $\leq m$", an equal number $t$ most frequent n-grams from each $1, 2, \cdots, m$ n-gram groups were selected so that they add up to the specified number of n-grams on the x-axis.

The results are shown in \autoref{fig:topN_exp}. For both datastores, we once again see that the accepted length generally increases as the number of n-grams increases, though at a decreasing rate. However, this is not always true as the accepted length peaks at 2.51\footnote{Note that this accepted length is only over the first 20 conversations, which is why it's higher than the accepted lengths reported in \autoref{sec:comparison-with-rest}.} for the Stack dataset at about 1.8M n-grams and then goes down. We believe the reason for this is the same reason why CREST can outperform REST: keeping token trees that are not common enough results in ``lower quality" or smaller token trees being used, resulting in a lower accepted length.

Interestingly, for both datasets, the best value of $m$ is not 2, even though the best value of $n$ was 2 in \autoref{fig:individual_n_exp}. We believe this is because even though 2-grams are generally better than 3-grams, as shown by \autoref{fig:individual_n_exp}, \textit{more common} 3-grams can still be better than \textit{less common} 2-grams. This motivates the need to compare n-gram frequencies across different values of $n$.

One difference between the two datasets is that the value of $m$ matters much less for ShareGPT than it does for Stack. We believe this is related to how the accepted lengths for ShareGPT in \autoref{fig:individual_n_exp} drop off much faster for higher values of $n$ than for Stack. As it's less valuable to have larger n-grams for ShareGPT, we don't see much of a benefit of increasing $m$ in \autoref{fig:topN_exp}.

\subsection{Token Tree Size Analysis}
\label{sec:avg_tree_size}
In order to focus our project on only studying a single technique and to have a fair comparison with REST, we retained the default hyperparameters recommended by REST. One of these hyperparameters is the number of tokens in the token tree, which REST set to 64. However, we later discovered that the size of token trees in practice was far smaller than 64, as seen in \autoref{tab:tree_size}. Because token trees tend to get smaller as n-grams get less frequent, the size of the datastore needs to be controlled. We chose to use the same datastore sizes as the largest datastores we built in \autoref{sec:single-n} for each value of $n$.

\begin{table}[t]
\caption{Average tree size (\# of tokens) of 1GB databases.}
\label{tab:tree_size}
\vskip 0.15in
\begin{center}
\begin{small}
\begin{sc}
\begin{tabular}{lcccr}
\toprule
Dataset & $n$ & \# Freq. N-Grams & Avg. \# Tokens \\
\midrule
ShareGPT & 1 & 28598 & 30.36 \\
ShareGPT & 2 & 1643949 & 20.68 \\
ShareGPT & 3 & 1839587 & 21.73 \\
ShareGPT & 4 & 2064567 & 19.34 \\
ShareGPT & 5 & 2171748 & 16.98 \\
Stack & 1 & 29777 & 31.92 \\
Stack & 2 & 1643946 & 23.49 \\
Stack & 3 & 1839587 & 25.76 \\
Stack & 4 & 2064567 & 24.42 \\
Stack & 5 & 2171748 & 23.12 \\
\bottomrule
\end{tabular}
\end{sc}
\end{small}
\end{center}
\vskip -0.1in
\end{table}

We observe tree sizes much smaller than 64 tokens because we do not have enough occurrences of most n-grams to build a full 64-token tree with the datasets we are using. This shows that there is a lot of room for improvement for a 64-token verifier LLM because we are not ``saturating" its verification capacity.

CREST becomes more attractive under this observation. When adding more data to REST, the storage size grows linearly (see \autoref{sec:rest_datastore_problems}). However, it is possible to add more data to CREST while \textit{maintaining the same number of n-grams in the database} and only increasing the size of the token trees. Essentially, we are not wasting space storing redundant copies of common n-grams, nor are we wasting space storing rare n-grams that would seldom be matched.

\section{Conclusion}
\label{sec:conclusion}
In this paper, we introduced CREST as an optimized version of REST. The key innovation behind CREST lies in the concept of ``uncoupling" n-grams and their continuations into a dictionary structure, which enables the targeted removal of specific n-grams. We then use the ``top $t$" method to select a subset of the ``smallest, most common" n-grams to include in the datastore.

Our experimental results demonstrate that CREST outperforms REST in terms of average accepted length across all tested datastore sizes. CREST matches REST's accepted token length with 10.6-13.5x less storage space and achieves a 16.5-17.1\% higher acceptance length across the HumanEval and MT Bench benchmarks. Overall, CREST is a robust architecture for retrieval-based speculative decoding that remains compact and effective even as the size of the pre-training dataset scales up.

\bibliography{min_ref}
\bibliographystyle{mlsys2021}






\end{document}